\PassOptionsToPackage{pdfa,pagebackref}{hyperref}
\documentclass[sigconf]{acmart}
\hypersetup{pdfapart=2, pdfaconformance=u}

\usepackage{enumitem}
\usepackage{booktabs}
\usepackage{caption}
\usepackage{bm}
\usepackage{subcaption}

\usepackage{siunitx}
\usepackage{caption}
\usepackage{subcaption}
\usepackage{tabularx}
\usepackage{multirow}

\newcommand{\x}{\bm{x}} 
\newcommand{\y}{\bm{y}} 
\newcommand{\bounds}{\bm{b}} 
\newcommand{\objectness}{\bm{o}} 
 
\newcommand{\regionmask}{\bm{r}} 

\newcommand{\region}{\bm{\hat{r}}} 

\newcommand{\mask}{\bm{m}} 
\newcommand{\disk}{\bm{m}^{\text{disk}}} 
\newcommand{\OSM}{\bm{m}^{\text{OSM}}} 
 
\newcommand{\X}{X}
\newcommand{\Y}{Y}
\newcommand{\trainset}{T}
\newcommand{\model}{f}
\newcommand{\Reals}{\mathbb{R}}
\newcommand{\iwidth}{w}
\newcommand{\iheight}{h}
\newcommand{\ichannel}{c}

\newcommand{\loss}{\mathcal{L}}
\newcommand{\suploss}{\mathcal{L}^{\text{sup}}}
\newcommand{\objloss}{\mathcal{L}^{\text{obj}}}
\newcommand{\npattern}{n}

\def\ie{\emph{i.e.}}

\def\etal{\emph{et al.}}

\copyrightyear{2023}
\acmYear{2023}
\setcopyright{acmlicensed}\acmConference[UrbanAI '23]{1st ACM SIGSPATIAL International Workshop on Advances in Urban-AI}{November 13, 2023}{Hamburg, Germany}
\acmBooktitle{1st ACM SIGSPATIAL International Workshop on Advances in Urban-AI (UrbanAI '23), November 13, 2023, Hamburg, Germany}
\acmPrice{15.00}
\acmDOI{10.1145/3615900.3628791}
\acmISBN{979-8-4007-0362-1/23/11}
\begin{document}

\title{Predicting urban tree cover from incomplete point labels and limited background information}


\author{Hui Zhang\orcid{0000-0002-0992-5830}, Ankit Kariryaa\orcid{0000-0001-9284-7847}, Venkanna Babu Guthula, Christian Igel\orcid{0000-0003-2868-0856}, Stefan Oehmcke\orcid{0000-0002-0240-1559}}
\email{{huzh,ak,vegu,igel,stefan.oehmcke}@di.ku.dk}

\affiliation{%
  \institution{University of Copenhagen}
  \department{Dept.~of Computer Science}
  \city{Copenhagen}
  \country{Denmark}
}


\begin{abstract}
Trees inside cities are important for the urban microclimate, contributing positively to the physical and mental health of the urban dwellers.
Despite their importance, often only limited information about city trees is available. Therefore in this paper, we propose a method for mapping urban trees in high-resolution aerial imagery using limited datasets and deep learning. Deep learning has become best-practice for this task, however, existing approaches rely on large and accurately labelled training datasets, which can be difficult and expensive to obtain.
However, often noisy and incomplete data may be available that can be combined and utilized to solve more difficult tasks than those datasets were intended for. 

This paper studies how to combine accurate point labels of urban trees along streets with crowd-sourced annotations from an open geographic database to delineate city trees in remote sensing images, a task which is challenging even for humans.
To that end, we perform semantic segmentation of very high resolution aerial imagery using a fully convolutional neural network.

The main challenge is that our segmentation maps are sparsely annotated and incomplete.
Small areas around the point labels of the street trees coming from official and crowd-sourced data are marked as foreground class.
Crowd-sourced annotations of streets, buildings, etc.{} define the background class. 
Since the tree data is incomplete, we introduce a masking to avoid class confusion.

Our experiments in Hamburg, Germany, showed that 
the system is able to produce tree cover maps, not limited to trees along streets, without providing tree delineations.
We evaluated the method on manually labelled trees and show that performance drastically deteriorates if the open geographic database is not used.

\end{abstract}



\keywords{urban environment, deep learning, point labels, sparse labels, crowd-sourced datasets, tree mapping, CNN, tree cover }


\maketitle

\section{Introduction}
\begin{figure}[t]
    \centering
    \includegraphics[width=\linewidth, clip]{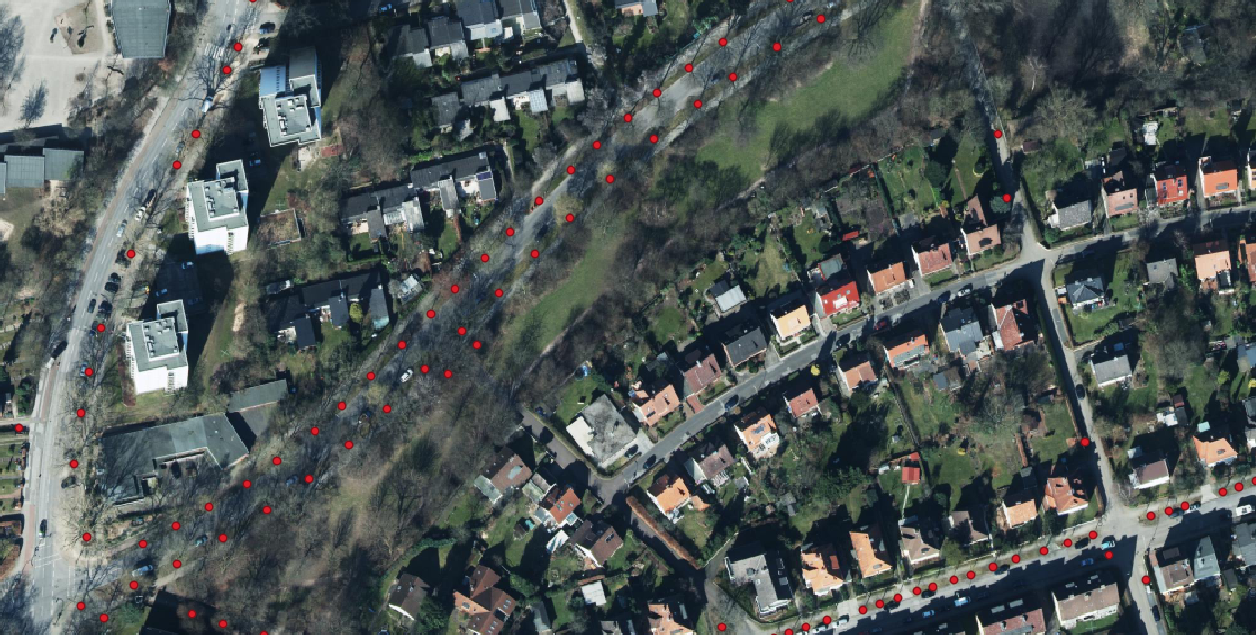}
    \caption{Aerial image from Hamburg, Germany with street trees overlaid in red. Street trees dataset is incomplete since it does not contain any information about the trees on private land, public parks, forests, or farms. }
    \label{fig:hamburg-trees}

\end{figure}

Trees are a vital component of our ecosystems.
They are vital for sustaining the biodiversity of various lifeforms and provide important services such as food, shelter, and shade \cite{aerts2011forest}.
In an urban setting, trees also offer benefits for physical and mental health~\cite{wolf2020urban}.
However, trees are also highly vulnerable to change in climatic conditions~\cite{bastin2019global}. 
Increase in global temperatures is associated with an increased global tree mortality rate, which reduces the ecosystem functioning and impacts their role in carbon storage~\cite{portner2022climate}. 
Monitoring the amount of trees is therefore vital to devise mitigation and adaptation measures against climate change.



In this paper, we propose a method to train a deep learning model to predict tree cover in an urban setting with sparse and incomplete labels.
This work differs from existing studies in several key aspects.
First, our work is unique in combining several different open data sources. 
To the best of our knowledge, no previous study has evaluated the potential of authority-managed tree records and crowd-source annotations from an open geographic database for tree mapping. 
Second, we focus on urban areas, which are relatively under-explored in other work~\cite{tucker:23,li2016deep}, although many free data sources exist.
Third, existing work relies on strong preprocessing and fully annotated data in which the object has either been accurately delineated~\cite{martins2021semantic, brandt2020unexpectedly} or been annotated by a bounding box or at least a point label. 
An example of point labels is done by Ventura \etal~\cite{ventura2022individual}, who manually annotated \num{100000} trees from eight cities in the USA and collected multiple years of imagery. 
Also, Beery \etal~\cite{beery2022auto} incorporated different sources of public data sets, but required multiple steps of data cleaning, resulting in nearly half of the tree records being removed.
In contrast, we exclusively utilize freely available data, both for input imagery and labels, which requires no annotation efforts for training. 

It is important to acknowledge that combining different sources of public data presents unique challenges, such as imbalanced classes and noisy labels, given that these data are not originally designed to be used together (see Fig.\ref{fig:hamburg-trees}). 
To make full use of the incomplete and sparsely labeled tree data as well as reduce the uncertainty of the background class, we proposed a mask regime that carefully selects pixels of trees and background with high probability of being that class. 
With this mask regime, we show that our approach is able to utilize this newly conjuncted dataset to predict urban trees with a balanced accuracy of 82\% on sparsely labeled data and 84\% on fully annotated data. 
We also introduce an objectness prior in the loss function inspired by weak supervision literature.
Originally proposed in \cite{objectnessPointCNN}, pseudo-labels are derived from model predictions that are pretrained on another dataset with the same task.
We derive pseudo-labels from an adapted watershed algorithm \cite{engineeredObjectness} to increase the extent of the object being sensed by the model for point-level supervision without requiring a pretrained model on the same task. 
Unlike common weak supervision scenarios that assumes sparse but fully annotated data, our incomplete annotations can lead to an incorrect objectness prior.
Consequently, we also applied our mask regime to the objectness prior and restrict learning of the target class to the area close to our tree labels.
This ablation study showed that the masking regime is always benifical, while the inclusion of an objectness prior is highly dependent on its quality.

To summarize, our main contributions are as follows:
\begin{itemize}
    \item A novel masking approach for combining noisy crowd sourced data with precise point labels.
    \item A dataset created from publicly available data, bringing forward the challenge of incomplete and sparse labels as well as a hand-delineated test set.
    \item An evaluation and comparison of different techniques to include the novel masking scheme.
\end{itemize}


\section{Related Work}


As in many other fields, deep learning learning models have become state-of-the-art method for mapping trees and tree cover in aerial, satellite and LiDAR imagery. 
However, training these models in a supervised learning setting requires large volumes of manually annotated data, which is often tedious and expensive to create and requires domain expertise. 
Typically, these methods are trained with dense labels, such as full delineations of trees. 
For example, ~\cite{brandt2020unexpectedly} manually annotated \num{89899} trees on very high-resolution satellite imagery for training their deep learning models. 

Recent research shows that semi- and weakly supervised learning have made great progress in the semantic segmentation of images~\cite{weaklyReview}. 
Weakly supervised learning aims to learn from a limited amount of labels in comparison to the entire image~\cite{weaklyReview, 5540060, YU201882}. 
Other works distinguish between different levels of weakly supervised annotations, such as bounding boxes~\cite{7410548}, scribbles~\cite{ScribbleSup}, points~\cite{engineeredObjectness, Zhang_2023_CVPR}, image labels~\cite{papandreou2015weakly}, pixel-level pseudo labels generated with class activation maps~\cite{Kang_2023_CVPR, Ru_2023_CVPR, Rong_2023_CVPR}, and also a text-driven semantic segmentation~\cite{Lin_2023_CVPR}. 
While fully-labelled data is limited, point labels are also used in instance segmentation methods, such as \cite{kim2023devil} introduced a novel learning scheme in instance segmentation with point labels and \cite{proposalInstancePointSup} proposed point-level instance segmentation with two branch network such as localisation and embedding branch.

Interactive segmentation with point labels started a few decades back and is still an active research topic ~\cite{937505}.
These segmentation models started training with point labels that annotate entire objects. 
Lin \etal~\cite{ScribbleSup} proposed ScribbleSup based on a graphical model that jointly propagates information from scribble and points to unmarked pixels and learns network parameters without a well-defined shape. 
Maninis \etal~\cite{extremPointCNN} proposed a framework with a point-level annotation that follows specific labeling instructions such as left-most, right-most, top, and bottom pixels of segments. 
Bearman \etal~\cite{objectnessPointCNN} proposed a methodology by incorporating objectness potential in the training loss function in segmentation models with image and point-level annotations. 
Li \etal~\cite{engineeredObjectness} utilised an objectness prior similar to \cite{objectnessPointCNN} but instead of a convolutional neural network (CNN) output they utilize distances in the pixel and colour space, meaning that the further away in the image and the more different the colour, the objectness decreases. 
Zhang \etal~\cite{Zhang_2023_CVPR} proposed a contrast-based variational model~\cite{Mumford1989OptimalAB} for semantic segmentation that supports reliable complementary supervision to train a model for histopathology images. 
Their proposed method incorporates the correlation between each location in the image and annotations of in-target and out-of-target classes. 
The weak supervision part of our research is inspired by \cite{objectnessPointCNN,engineeredObjectness,Zhang_2023_CVPR}, as we have only a single point for each tree, we use point labels in combination with denser background information while considering an objectness prior.

In contrast to these scenarios, we consider the added challenge of incomplete annotations, meaning some relevant objects in an image might not be annotated at all.


\section{OpenCityTrees Dataset}\label{Dataset}
Public agencies often maintain valuable records of trees and other public attributes such as roads, parking areas, buildings, etc.
These datasets are, however, often noisy due to differences in collection techniques, lack of the common data collection standards, noisy sensors, and lack of records of temporal changes. 
Moreover, they are mostly not developed for the goal of training supervised deep learning models or for use in conjunction with other modalities such as aerial or satellite imagery.
As such, they are potentially underutilized in research.
To demonstrate their usefulness, we created a new dataset for weakly supervised segmentation from such records.\footnote{https://doi.org/10.17894/ucph.b1aa4ca2-9a4b-40d0-aa87-c760e69bf703}
\subsection{Input images}

To demonstrate the usefulness of public but incomplete datasets, we use the aerial images from Hamburg, Germany as input for our models. 
These images contain 3 channels (RGB) at a \SI{0.2}{\metre/pixel} resolution and they were downloaded from the data portal of the Spatial Data Infrastructure Germany (SDI Germany)\footnote{\label{geop}\url{https://www.geoportal.de/portal/main/}}.
The images were captured in May 2016. 
As seen in Fig.~\ref{fig:hamburg-trees}, the individual features such as trees, buildings, and cars on the streets are visible to human eyes. 
We downloaded 27 image tiles of $5000\times5000$ pixels (\ie~covering a $\SI{1}{\kilo\metre} \times \SI{1}{\kilo\metre}$ area ) within the bounds (9.9479070E, 53.4161744N,  9.9684731E, 53.6589539N). 
These tiles extend from the north to the south border of Hamburg but are limited to \SI{1}{\kilo\metre} strip close to the city center. 
Hamburg is situated on the coast of the Elbe river with a densely populated city-center. 
Along it's border (\ie~away from the city center), the city-state also contains suburbs, farms and forested area. 
The chosen images capture all these different characteristics of the city along the north-south gradient. 
Since images are captured in early spring, many of the trees are without leaves, making certain trees more challenging to identify.

When designing the dataset, we considered that additional height data derived from LiDAR could potentially enhance results.
However, we decided against including it because there are practical constraints associated with LiDAR data availability and collection. 
High-resolution LiDAR data (e.g., submeter similar to our RGB source) often remains inaccessible due to regulatory limitations, especially concerning drone or plane flights over urban areas. 
Additionally, acquiring LiDAR measurements is more costly compared to RGB measurements, which could make frequent temporal analyses of urban tree cover infeasible. 
For instance, the open data portal we used, does not have submeter height measurements available for Hamburg.

\subsection{Label data}

Two sources of labeled data are combined:

\paragraph{Ground truth for trees}
The Authority for Environment and Energy of the city of Hamburg maintains a list of all street trees\footnote{\url{http://suche.transparenz.hamburg.de/dataset/strassenbaumkataster-hamburg7}} as recorded on the 6th of January 2017.
The dataset contains various attributes of individual trees such as location, height, width, species, age, and condition. 
However, as the name suggests, this information is limited to the trees along the streets of Hamburg and does not include information about trees on private land, in public parks, or in forested areas. 
Unlike other data usually used in point-supervision where each object is assumed to be annotated with at least one point, we have incomplete annotations, increasing the ambiguity of the background class. 
In the area of interest, the dataset contains information about \num{11366} trees. 
These trees are from \num{136} unique species. 
In Fig.~\ref{fig:hamburg-trees}, trees in the street trees dataset are overlaid in red circles. 
Each tree is provided as a point referenced in a local reference system (EPSG:32632 - WGS 84).
However, the point location of the tree label can be inaccurate, for example, it might not overlap with the center of the tree or, in the worst case, any part of the trees due to the geo-location errors.
Another challenge with the dataset is that distribution of species of the street trees may vary significantly from the distribution of trees species in forests, parks, farms, or gardens.  

As a second source of ground truth data, we use OpenStreetMap (OSM)~\cite{haklay2008openstreetmap}. 
Within these bounds and the tag {'natural':'tree'}, OSM contains the location of \num{6375} trees. 
Out of these, \num{145} trees contain species information (\num{24} unique species).
These OSM data offer information regarding trees in private and public areas along with street trees. 

\paragraph{Ground truth for non-trees classes }

\begin{table}[ht]
\centering 
\caption{Description of objects defining the non-tree class. The buffer distance is in meters and negative buffers shrinks the object. Only vectors with non-negative area were chosen.}
\setlength{\tabcolsep}{1.pt}
\begin{tabularx}{\linewidth}{>{\hsize=.2\hsize}XlS[table-format=5.0]S[table-format=1.0,table-space-text-post = \si{\meter}]>{\hsize=.5\hsize}X}
\toprule
\textbf{Type} & \textbf{OSM tag} & \textbf{Count} & \textbf{Buffer} & \textbf{Comments} \\ \midrule
Buildings   &  {'building':True}  &   23075   &   -5\si{\meter}        &  Buildings of all types  \\ \addlinespace
Roads   &  {'highway':True}     &   111    &    -7\si{\meter}        &   Mostly around parking areas or bus terminals \\ \addlinespace 
Sports pitches     &    {'leisure':'pitch'}    &   135  &  -7\si{\meter}       &  Soccer pitches and similar types of grass surfaces \\ 
\bottomrule
\end{tabularx}%
\label{non-tree-classes}

\end{table}

Table~\ref{non-tree-classes} provides an overview of the objects that we use to define the non-tree class. 
The non-tree classes are mostly dominated by buildings which provide relevant information about different construction material and roof types.
While the area contains abundant roads, it is a tricky class to consider for the true negatives since the trees are often planted next to the roads and large parts of tree canopies overlap with roads. 
We only used road data if they had an associated area (i.e. stored as polygon or multi-polygon). We used OSMNX library to download data from OSM~\cite{boeing2017osmnx}.
Sports pitches, which includes grassed surfaces such as soccer pitches, are limited to 135 instances and it is only classes that provides information on grass which is easy to confuse with trees.

\subsection{Challenging aspects of the data}
By combining tree inventories and geographic data from existing public records, we create a rich dataset, without the need for additional acquisition of labor-intensive annotations. 
Public records maintain valuable information about trees and other public attributes. 
However, using incomplete public records for tree prediction also introduces a number of challenges:

\textbf{Sparse labels}: 
The ground truth of trees are given as point labels that cover most public streets, some public parks, and a few private places.
These annotations are incomplete and only represent a small portion of urban trees. 
In addition, these street trees are also sparsely distributed.

\textbf{Presence of noise}: 
Although the tree census data and aerial images are obtained from relatively close point of times, it is important to note that changes in the tree population might have occurred during the time gap. 
Trees could have been removed, died, or new trees might have been planted. Besides, there are geo-location errors as mentioned before, any nearby pixel of the tree could be labeled as the tree centroid.

\textbf{Image quality}: 
The quality of aerial imagery can vary for different tree species. 
The images were captured in early spring, when deciduous trees have not yet grown leaves. In addition, renewal of growth in trees near streets may be influenced by extended period of illumination and emissions from the streets~\cite{matzke1936effect}. 
As a result, these trees may not be well represented in the aerial image.

\textbf{Invisible trees}: 
There are trees located within shadows of nearby tall buildings, darkening the image and increasing potential class confusion between tree and shadows.

\begin{figure}[t]
    \centering
    \includegraphics[width=0.99\linewidth]{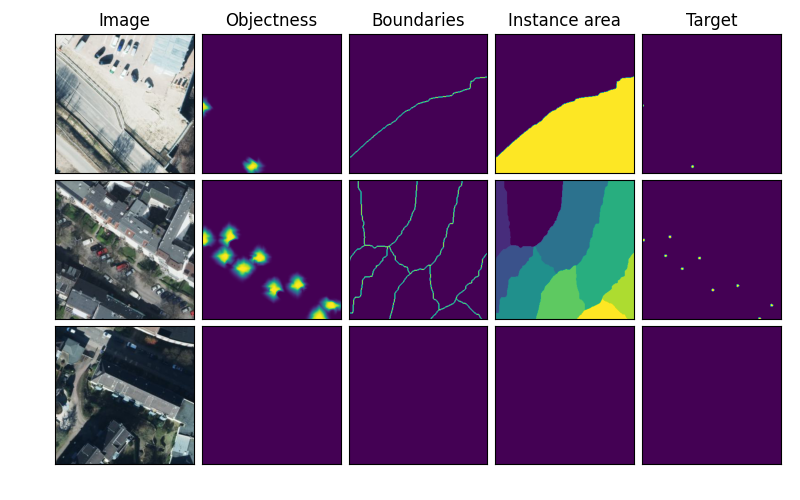}
    \caption{Objectness prior maps (col 2) and instance areas (col 4) were generated using input images (col 1) and locations of tree centroids (col 5) according to \cite{engineeredObjectness}. The boundaries (col 3) are derived where instance areas are touching. \label{fig:objectness-map}}

\end{figure}

\section{Learning from Incomplete \& Sparse Labels}

Our main challenge in training a tree segmentation model is obtaining accurate and effective labels.
Coming from open-data sources, however, labels are incomplete, meaning that not all trees or non-tree objects in an area are annotated.
These incomplete labeling deviates from the typical definition of weakly supervised learning~\cite{objectnessPointCNN,engineeredObjectness}, where we assume sparse labels (e.g., points, scribbles, bounding boxes, \ldots) are available for every relevant object.
In addition, tree labels are only available on a point level, meaning a single point represents a tree although the tree canopy encompasses a larger area. 
The non-tree labels are taken from OSM thus describing only parts of the image, in addition, we chose to shrink their shape to avoid overlap with potential tree that are not covered by our dataset (e.g., a tree reaching over a building), see Table~\ref{non-tree-classes}.

We frame the learning task as binary semantic segmentation of trees and introduce concepts to deal with the incomplete sparse labels. 
To that end, we consider a training set \[\trainset = \left\{(\x_1, \y_2), \ldots, (\x_\npattern, \y_\npattern)\right\} \subset \X \times \Y\] with images  \(\x_i \in \X = \Reals^{\iwidth\times\iheight\times\ichannel}\), a segmentation mask \(\y_i \in \Y = \{-1, 1\}^{\iwidth\times\iheight}\), and number of samples $\npattern$.
Further, $\iwidth, \iheight,$ and $\ichannel$ correspond to the width, height, and number of input channels, respectively.
The pixels containing the non-trees objects are considered as the negative class samples.
In our training dataset, we treat the pixels in a \SI{60}{\cm} radius ($7\times7$ pixels) around the point coordinate of a tree as positive class labels, which increases the number of positive training labels substantially. 

Training in such a setting is non-trivial.
For example, learning a semantic segmentation only given point labels is challenging because information about the spatial expand of the objects in question is limited.
Previous research introduces this spatial expand information by means of an objectness prior. The objectness prior gives an estimate of the class likelihood per pixel. 
As shown in Figure \ref{fig:objectness-map}, given the location of the trees, the algorithm estimates the potential spatial extent for each tree.

This prior can come from pretrained models on similar tasks \cite{objectnessPointCNN}, but also from classic algorithms, e.g. inspired by watershed segmentation \cite{engineeredObjectness}.
Our approach uses these two loss functions in conjunction:
\begin{align}
    \loss = &\suploss(\model(\x) \odot \mask, \y \odot \mask)~+ \nonumber \\
            & \objloss(\model(\x) \odot \regionmask, \objectness \odot \regionmask, \region \odot \regionmask) \cdot \beta ~,
\end{align}
where \(\suploss\) is the supervised loss (e.g., binary cross entropy (BCE) loss) that learns from the labeled data and the objectness loss \(\objloss\), where this prior information is utilized. 
Here, $\odot$ denotes the selection operator that chooses elements where the learning mask $\mask$ is set to 1 and returns the elements as a flattened vector.
The parameters of \(\objloss\) are the predictions \(\model(\x)\), the objectness prior~\(\objectness\), and an instance region~\(\region\).
Note, since no pre-trained CNN~\cite{objectnessPointCNN} on tree segmentation was easily available, we utilized the method described by \cite{engineeredObjectness} to calculate \(\objectness\) and \(\region\).
To obtain \(\objectness\), we calculate the distance matrix \(\Delta\in\Reals^{\iwidth, \iheight}\) by applying the adjusted watershed algorithm \cite{watershed} as in \cite{engineeredObjectness} with the point labels being used as markers and then transforming these distances into a pseudo-probability distribution \(\objectness = e^{-\alpha\Delta^2}\), with $\alpha=10$ to create fast decay of values the farther away from an actual label.
The current settings for these pseudo-probabilities were explored during a preliminary study on the training set but the ones provided by \cite{engineeredObjectness} turned out to perform best.
From the same adjusted watershed output, we use the watershed instance assignments as \(\region\).
\(\beta\) is trade-off parameters to change the influence of the objectness loss.
See Figure~\ref{fig:objectness-map} for an exemplary input and objectness-related attribute.
In our incomplete label setting, the generated objectness can only capture the trees indicated by point labels.
Therefore, to represent where labels are available, we declare two learning masks \(\mask \in \{0, 1\}^{\iwidth\times\iheight}\) and \(\regionmask \in \{0, 1\}^{\iwidth\times\iheight}\), where $1$ means a label is present and $0$ corresponds to missing label information.
These masks can be defined in several ways as we explore in the experimental section and can be considered one of main contributions of this paper.

For the objectness loss, we extend the binary cross-entropy similarly to \cite{engineeredObjectness,objectnessPointCNN}
\begin{equation}
\objloss = - \frac{1}{|\objectness|}\sum_{i=1}^{|\objectness|} \text{BCE}(\model(\x)\odot\region_i, \objectness\odot\region_i) ~,
\end{equation}
where each tree instance is calculating its own loss value depending on the instance region $\region \in \{0, 1\}^{|\objectness|\times\iwidth\times\iheight}$. 
The number of tree instance \(|\region|\) changes for each sample, as does the number of pixels in each region.
Averaging inside the instance sum effectively weights each instance the same, regardless of size.

\section{Experimental Evaluation in Hamburg}
\label{sec:exp}

\subsection{Ablation study}
\newcommand{\Mask}{\textit{Mask}}
\newcommand{\Obj}{\textit{Obj}}
\newcommand{\MaskObj}{\textit{MaskObj}}
\newcommand{\MaskObjThresh}{\textit{MaskObjThresh}}
\newcommand{\baseline}{\textit{baseline}}

The choice of masks \(\mask\) and \(\regionmask\) is crucial in our sparse and incomplete label setting.
To that end, we compared five different training scenarios as shown in Table~\ref{tab:ablation}.
The \baseline\ scenario is only using the supervised loss without any masking.
The public authority and OSM tree labels are expanded from a point to a disk \(\disk\) of radius \SI{1.5}{\metre}, which is indicated by \(\disk = 1\).
The second scenario, called \Obj, uses the objectness loss over the entire image in combination with supervised loss and we mask out all pixels with negative labels except on the boundaries of the instance region \(\region\) (see Figure~\ref{fig:objectness-map}). 
\Obj\ is a reimplementation of \cite{engineeredObjectness}.
The third scenario uses the supervised loss along with our proposed masking scheme, termed \Mask. 
Here we do not consider the objectness loss and only evaluate the supervised loss where we have positive labels (indicated by \(y=1\)), and where we have information about the shrunken OSM non-tree objects \(\OSM\), which is indicated by \(\OSM = 1\). 
In addition, in shrunken OSM non-tree objects, we remove negative pixels that are within \SI{1.5}{\metre} of a positive label.
In the fourth scenario, we combine our masking approach with objectness in \MaskObj, by employing the objectness loss but restricting it to the \SI{1.5}{\metre} radius around the positive labels.
Lastly, we add an additional constraint to the objectness by ignoring all the pseudo-probabilities that are below \num{0.2}, which will reduce the learning about the negative class in \(\objloss\), which we refer to as \MaskObjThresh.

\subsection{Network architecture, loss function, and hyperparameters}
To address the tree segmentation task, we employ a fully-convolutional network based on the U-Net architecture~\cite{unet}. 

\textbf{Experimental settings.} Among other things, in the past U-Nets have been used for semantic segmentation of trees in satellite imagery~\cite{brandt2020unexpectedly}.
We adapted the U-Net architecture by applying batch normalization~\cite{ioffe2015batch} instead of dropout layers and replacing ReLU with ELU~\cite{clevert2015fast} as activation functions. 
We use binary cross entropy (BCE) as loss function as our supervised error measure.
The aerial images were split into $300\times300$ patches and a batch consists of 36 patches.
For training and hyperparameter optimization, the dataset was split into 80\% training set (\num{3566} patches), 20\% validation set (\num{788} patches).
To improve training stability, we accumulated gradients over 14 batches (i.e 504 images) before the optimizer step. 
The model is trained for 500 epochs and the final weights were chosen w.r.t.\ the best recall score on the validation set. 

\textbf{Evaluation on sparse and dense labels.} 
The evaluation of the model's performance was done with two types of data annotations, point annotation, and dense object annotation. 
These two datasets are spatially independent.
First, we evaluated the model on the point-annotated data from 28 tiles (\num{4169} patches) within the bounds (9.962748E, 53.407065N,  9.83603E, 53.658832N), which is a \(\SI{1}{\kilo\metre}\times\SI{28}{\kilo\metre}\) stripe adjacent to the training data stripe.
None of these tiles were used for training or intra-model validation and the ground truth dataset for them was created in exactly the same way as described in Section~\ref{Dataset}. 
None of the pseudo labeling (e.g., extending of point labels to 4 pixels or a disk as label) was utilized during evaluation, meaning that for point labels only the corresponding pixel is considered and for the background class only the negatively buffered area.
The sparse street tree dataset and OSM had information on \num{14137} trees within these bounds. 

To evaluate our models performance on dense object prediction, we manually annotated a tile within 
a $\SI{1}{\kilo\meter^2}$ area
, which is \SI{3}{\kilo\metre} to the east of our training data.
The delineation work was done using QGIS and is mainly based on the input image, which was cross-referenced with Bing and Google Satellite Maps. 
The annotation was then verified within the authors' group, which eliminates some bias.
To utilize this dataset for an unbiased tree cover estimate, we split it further into a model selection set and a test set.
We applied only the best model from the model selection set in terms of IoU to the test set.

\subsection{Sparse Label Results}

\begin{figure}
    \centering
    \begin{subfigure}[b]{0.32\linewidth}
    \centering
    \includegraphics[width=\linewidth,trim={0 1cm 0 0},clip]{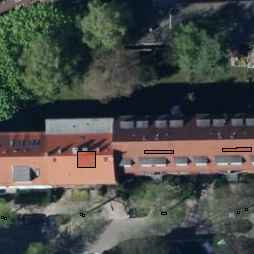}
    \includegraphics[width=\linewidth,trim={0 1cm 0 0},clip]{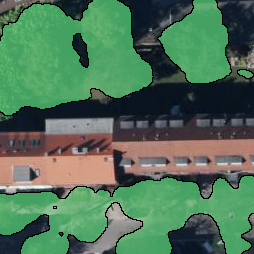}
    \caption{Example 1}
    \end{subfigure}
    \hfill
    \begin{subfigure}[b]{0.32\linewidth}
    \centering
    \includegraphics[width=\linewidth,trim={0 1.075cm 0 0},clip]{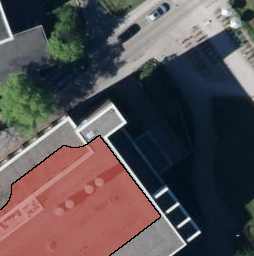}
    \includegraphics[width=\linewidth,trim={0 1.075cm 0 0},clip]{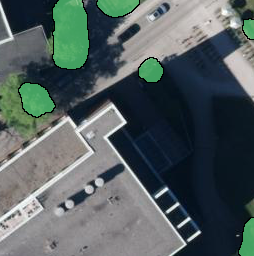}
    \caption{Example 2}
    \end{subfigure}
    \hfill
    \begin{subfigure}[b]{0.32\linewidth}
    \centering
    \includegraphics[width=\linewidth,trim={0 1cm 0 0},clip]{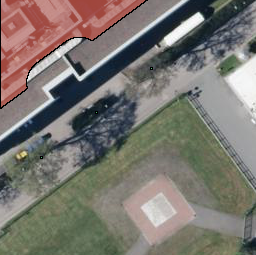}
    \includegraphics[width=\linewidth,trim={0 1cm 0 0},clip]{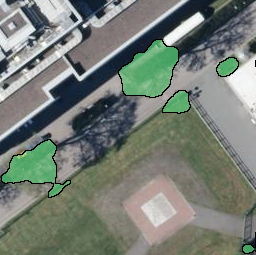}
    \caption{Example 3}
    \end{subfigure}
    \caption{Three examples of target (top) and possible predicted segmentation (bottom). The predicted positive class is overlayed in green. In the target examples, a red overlay indicates the negative class and transparency means the learning mask is 0. For the predicted segmentation, only the positive class is shown and the negative class is transparent. 
    } \label{results-segmentation}
\end{figure}

    




The results of the sparse label test set are given in Table~\ref{tab:ablation}.
It is crucial to acknowledge the highly imbalanced nature of the dataset when evaluating with sparse labels. 
Due to this significant imbalance, the number of false positives can be far greater than the true positive, leading to a substantially low precision value. 
Specifically, due to the class imbalance with \num{2448} times more negative class pixels than positive class pixels, the precision of our models was only around 3\%.
Therefore we focus on the recall (sensitivity) of the target class and balanced accuracy (BA) to evaluate the model performance on sparse labels.

The \baseline\ model performed worst and appears to mainly predict the background class.
Performing best was the \Mask\ model w.r.t. recall with \num{90}\% and \MaskObj\ w.r.t.\ BA with \num{84}\%.
Even though the BA of \Obj\ is close to the mask models with \num{78}\%, the recall value is comparatively low with \num{59}\%.
In Figure~\ref{results-segmentation} exemplary target and prediction segmentation masks are shown.

\begin{table*}
\centering
\caption{
Results on sparse and delineated data across different ablation settings.
For model selection set of the delineated labels, we compare intersection over union of the tree (IoU) of the tree class, F1, and balanced accuracy (BA) scores. 
The sparse labels are compared w.r.t.\ their recall and BA scores.
Additional masks are the shrunken OSM non-tree objects \(\OSM \in \{0, 1\}^{\iwidth\times\iheight} \), a \SI{1.5}{m} disk \(\disk \in \{0, 1\}^{\iwidth\times\iheight}\) around each positive values in \(y\), the bounds~\(\bounds\) between instances derived from the instance region map~$\regionmask$, and a mask of ones \(\bm{1}^{\iwidth\times\iheight} \in \{1\}^{\iwidth\times\iheight}\).
Results of the \baseline\ model were not calculated for the delineated data set since the model was discarded due to the sparse label performance.
\label{tab:ablation}}
\begin{tabularx}{\linewidth}{@{\extracolsep{\fill}}*{9}{l}}
\toprule
Name & Description & Setting & \multicolumn{3}{c}{Delineated}& \multicolumn{2}{c}{Sparse}\\
\cline{4-6}\cline{7-9}
 & & & $\text{IoU}_{\text{d}}$ & $\text{F1}_{\text{d}}$ & $\text{BA}_{\text{d}}$ & $\text{Recall}_{\text{s}}$  & $\text{BA}_{\text{s}}$\\
\midrule
\baseline&\multirow{3}{*}{\parbox{4cm}{Neither masking nor objectness and enlarging positive labels to a circle with \SI{1.5}{m} radius.}}&  $\beta = 0$ && &&&\\
&&                                                                                                                      $\mask = \bm{1}^{\iwidth\times\iheight}$ & \multicolumn{1}{c}{---} & \multicolumn{1}{c}{---} & \multicolumn{1}{c}{---} &0.0005 & 0.5002\\
&&                                                                                                                      $\y = \disk$ &  &&&& \\
\addlinespace
\cline{3-3}
\addlinespace
\Obj&\multirow{1}{*}{\parbox{4cm}{Reimplementation of \cite{engineeredObjectness}.}}&   $\beta = 1$  & & &&&\\
&&                                                                                      $\mask=\y \cup \bounds$ &  0.1191& 0.5479 & 0.5575  & 0.5908 & 0.7844\\
&&                                                                                      $\region = \bm{1}^{\iwidth\times\iheight}$  & & & &&\\
\addlinespace
\cline{3-3}
\addlinespace
\Mask &\multirow{2}{*}{\parbox{4cm}{Only supervised loss and masking out unknown areas.}}&  $\beta = 0$  &&&&&\\
(\textbf{ours})&&                                                                           $\mask = \y \cup (\OSM \setminus \disk)$& \textbf{0.4839}  & 0.7551 & \textbf{0.8119} & \textbf{0.8994} &0.8205 \\
\addlinespace
\cline{3-3}
\addlinespace
\MaskObj&\multirow{3}{*}{\parbox{4cm}{As \cite{engineeredObjectness} but restricting objectness loss to \SI{1.5}{m} radius around points.}}&   $\beta=1$& & &&&\\
(\textbf{ours})&&                                                                                                                           $\region = \disk$ & 0.3364& 0.7289&0.6978 &0.7771 &\textbf{0.8399}  \\
&&                                                                                                                                          $\mask = \y \cup (\OSM \setminus \disk)$ & & &&&\\
\addlinespace
\cline{3-3}
\MaskObjThresh&\multirow{3}{*}{\parbox{4cm}{As MaskObj but removing objectness smaller than the specified threshold (\(t=0.2\)).}}& $\beta=1$  & & &&&\\
(\textbf{ours})&&                                                                                               $\region = \disk \cap (\objectness \geq t)$ & 0.4805 & \textbf{0.7660} & 0.7870 & 0.8345 & 0.8135&\\
&&                                                                                                              $\mask = \y \cup (\OSM \setminus \disk)$ & & &&&\\
\bottomrule
\end{tabularx}
\end{table*}


\begin{figure}
    \centering
    \begin{subfigure}[b]{0.5\linewidth}
    \centering
    \includegraphics[width=\linewidth,trim={0mm 2.2cm 0mm 0cm},clip]{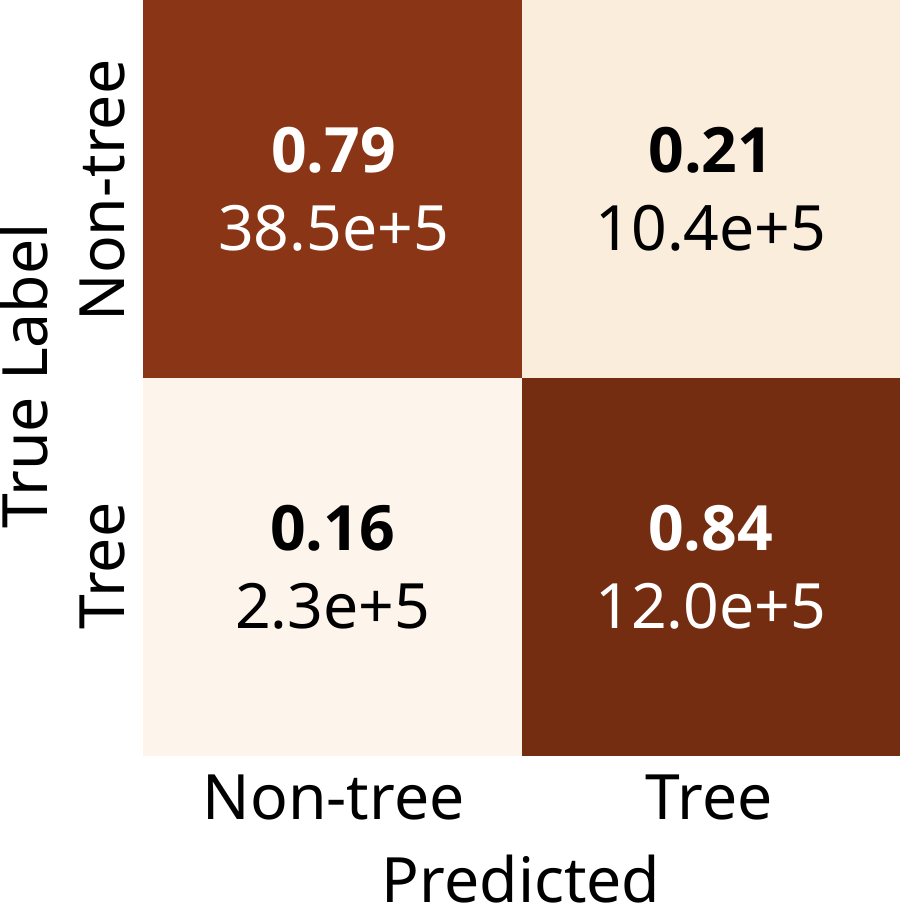}
    \caption{\Mask}
    \end{subfigure}
    \hfill
    \begin{subfigure}[b]{0.4275\linewidth}
    \centering 
    \includegraphics[width=\linewidth,trim={2.2cm 2.2cm 0mm 0cm},clip]{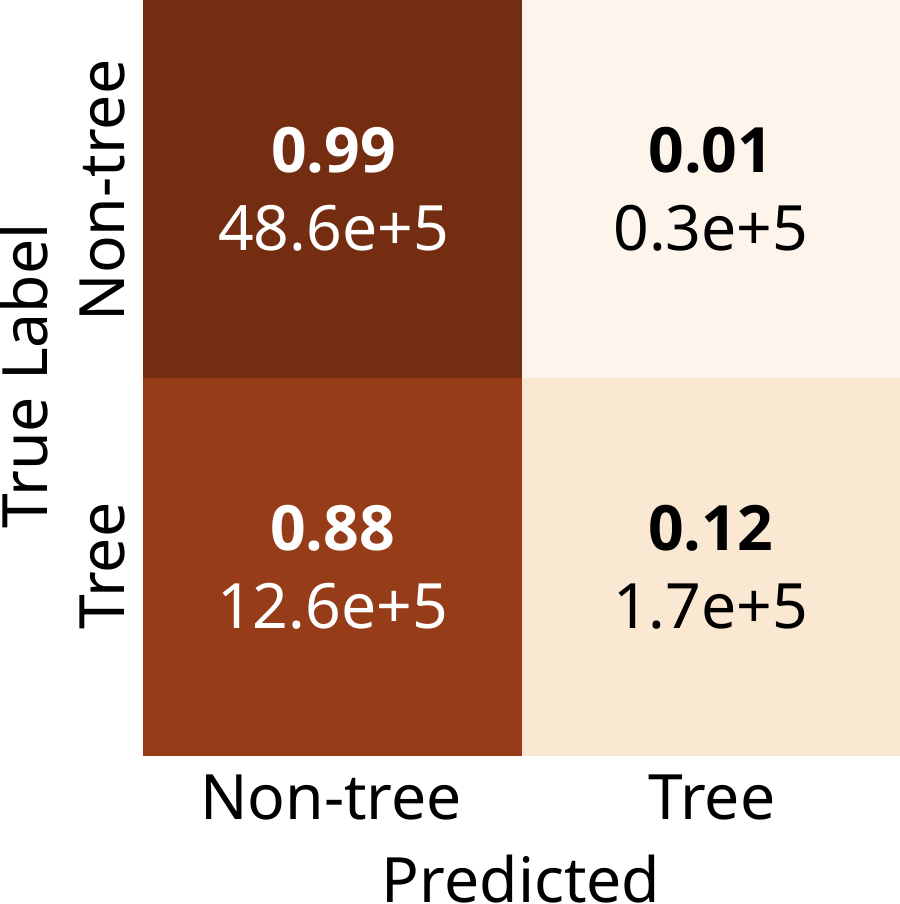}
    \caption{\Obj}
    \end{subfigure}
    \begin{subfigure}[b]{0.5\linewidth}
    \centering
    \includegraphics[width=\linewidth,trim={0cm 0.0cm 0mm 0cm},clip]{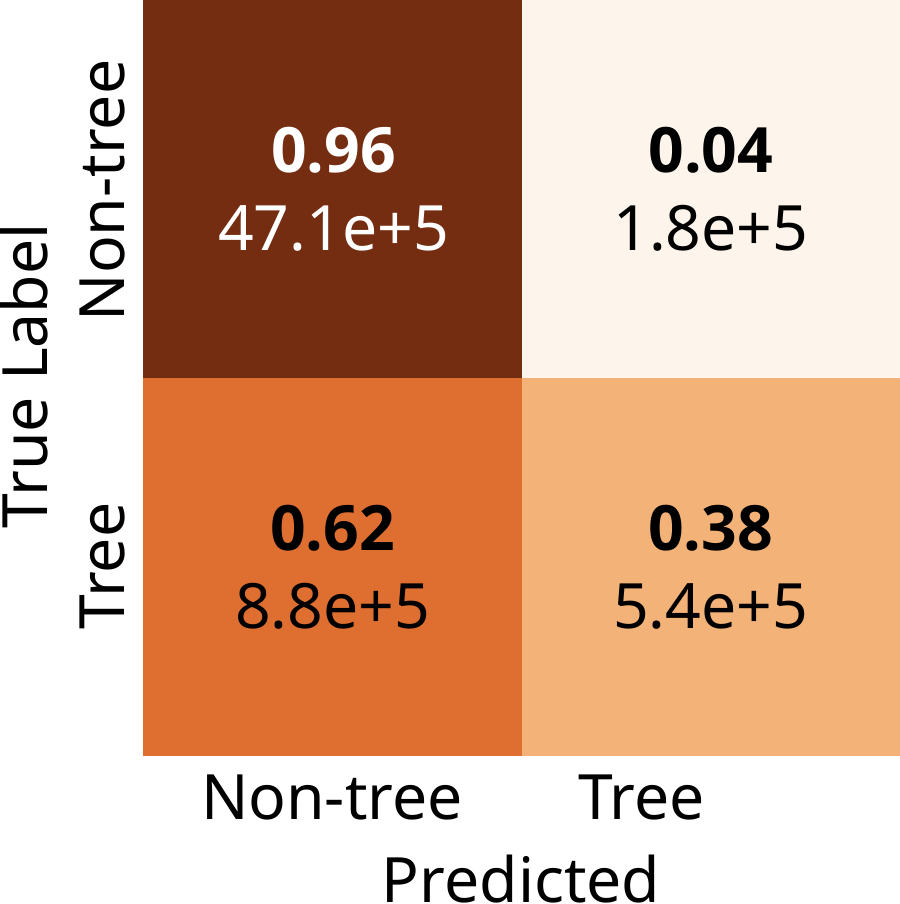}
    \caption{\MaskObj}
    \end{subfigure}
    \hfill
    \begin{subfigure}[b]{0.4275\linewidth}
    \centering
    \includegraphics[width=\linewidth,trim={2.2cm 0mm 0mm 0cm},clip]{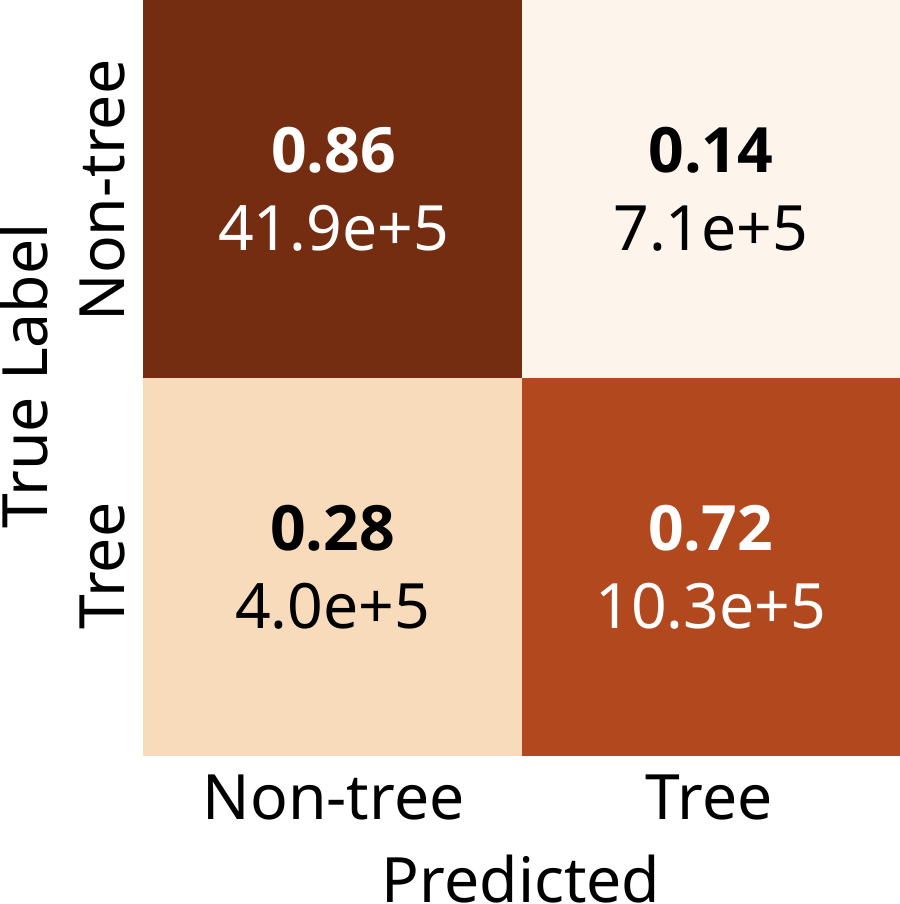}
    \caption{\MaskObjThresh}
    \end{subfigure}
    \caption{Confusion matrix across different ablation settings, created from the delineated model selection dataset. The normalized confusion matrix is shown in bold font (top number) and through color, while the lower number represents absolute number of samples.
    (a) \Mask: Our initial method that utilizes masking,
    (b) \Obj: Model as presented in \cite{engineeredObjectness}, 
    (c) \MaskObj: Combing masking and objectness, 
    (d) \MaskObjThresh: Restricting learning to positive labels for objectness loss. 
    We do not show the \baseline\ model here, because the model mainly predicted the negative class (e.g., in the first column both values are close to 1). 
    } \label{fig:ncmat}
\end{figure}



\subsection{Delineation Results}

\begin{figure*}
    \centering
    \includegraphics[width=\linewidth]{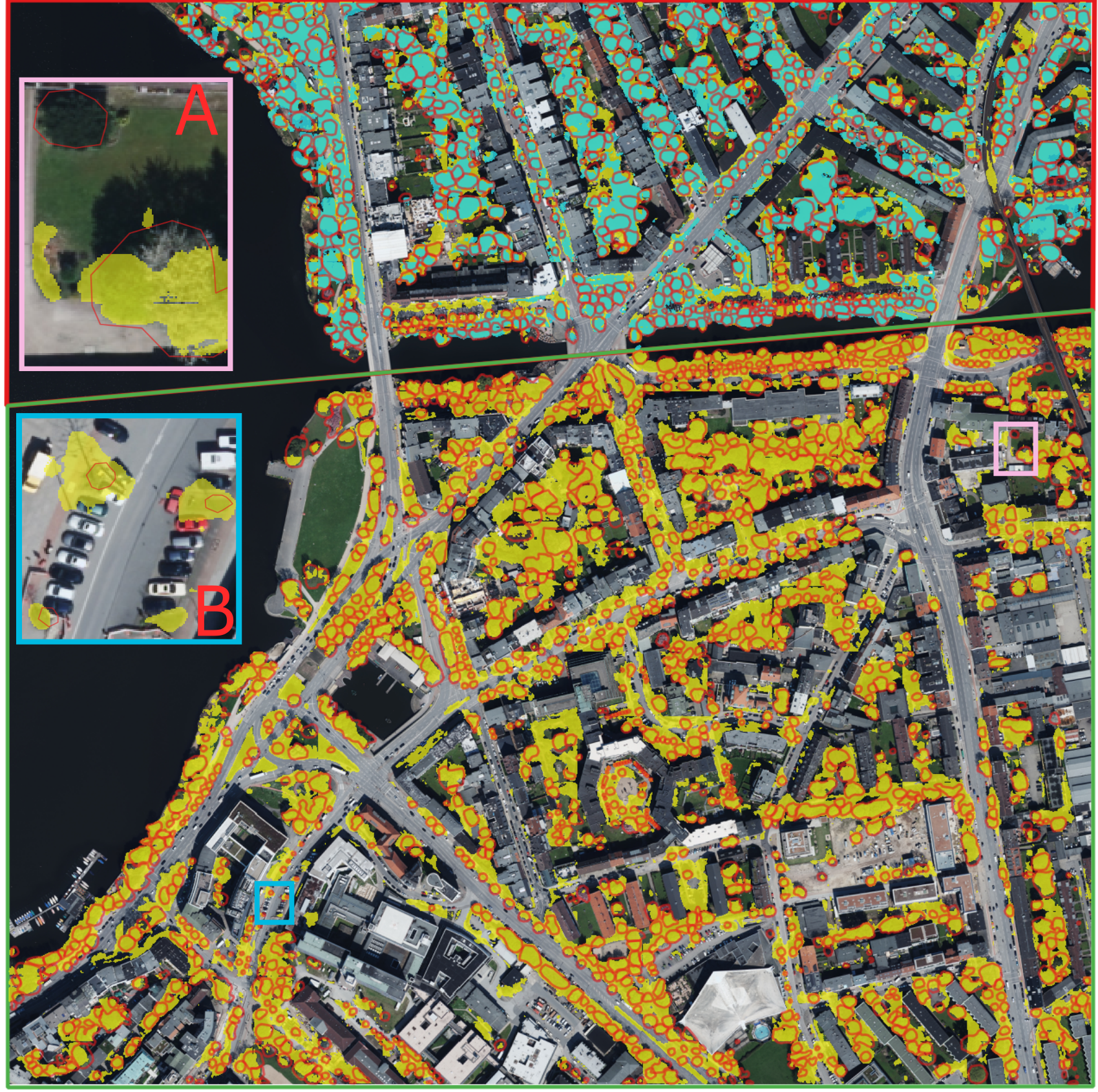}
    \caption{Results on the fully delineated dataset. Red polygons are the manually annotated ground truth. Validation data is shown within the red bounds, which is used for the model selection. As discussed in Section~\ref{sec:exp}, the \Mask~model is the final model chosen for prediction, which shows as the yellow mask on top of the satellite image. The segmentation mask in green are from the model \MaskObjThresh. The area bounded by the green lines shows the test set. Within the test area, we only apply \Mask.
    In cutout A the model predicts one of the trees correctly and interestingly does not predict the shadow of said tree. This tree is also of bright color since it is in bloom making it a hard example. However, the second tree in top left corner was not found, possibly because the coloration is close to the one of shadows. After manual rechecking of the predicted tree in left lower corner, we found that there is a tree not annotated originally (e.g., a false negative label).
    Cutout B is interesting because it contains trees without leafs. The model annotated these trees effectively and even finds another tree that was missed by the annotators (bottom right). This also demonstrates how difficult it is to define clear boundaries for these trees.
    \label{fig:delineation results} }
    
\end{figure*}

The intersection over union (IoU), F1, and BA score on the model selection set can be seen in Table~\ref{tab:ablation}.
We omitted the \baseline\ because of the subpar results on the sparse test set.
The class imbalance changes since these annotation are fully delineated, particularly, we now have \num{3.42} negative pixels for one positive pixel.
This change makes the use precision viable, which is why we consider the F1-score.
The original masking model \Mask\ performed best in IoU and BA, even though the difference in IoU compared to \MaskObjThresh\ is marginal.
\Obj\ only shows a BA of 56\% which is much smaller than the 78\% on the sparse data, showing a lack of generalization for this approach.
Using any kind of masking scheme seems to improve the results.

In Figure~\ref{fig:ncmat} we show the normalized and unnormalized confusion matrix results on the model selection set.
Note that \Obj\ and \MaskObj\ underpredicts the positive class, which corresponds to the low accuracy and a false-negative rate.
For \MaskObjThresh\ and \Mask\ the accuracy of the positive class is considerably higher, even though the false positive rate increases.
Interestingly, thresholding the objectness prior improves performance compared to the model without, indicating that the prior hold misleading information regarding the background class.

Based on the results on the model selection set, we decided that the best model is the one without any objectness loss but with masking (\Mask).
This decision is based on the better metrics but also on the simpler learning setup.
Our masking regime only requires a sensible choice for the excluded areas, while the objectness prior additionally requires a choice of how to calculate it.
For delineated test set the \Mask\ model achieved an IoU of \num{0.4253}, F1 score of \num{0.7393} and BA of \num{83.63}\%.
These results are comparable to the other sets, indicating a good generalization performance.

Finally, the predictions on the dense label set are shown in Figure~\ref{fig:delineation results}.
Within the bounds of the test area (\SI{591609}{\metre^2}), we detected a total tree cover of \SI{177142}{\metre^2} (\num{29.9}\%) compared to the annotated \SI{96946}{\metre^2} (\num{16.4}\%).
This shows an overestimation but as seen in the map, some of the trees were missed in the annotated dataset or were difficult to delineate without ambiguity.

\subsection{Discussion of Ablation results}
\textbf{Baseline.} 
The baseline segmentation model trained on incompletely labeled data with ambiguous information about the target class and the background class is not able to learn tree features thus predicts all pixels as background. 
As shown in the first row of Table~\ref{tab:ablation}, the recall of the target class is close to 0 when evaluated on the sparse labels. 
The model mainly predicts the background class, which aligns with our assumption that the class imbalance without our masking is challenging to overcome.

\textbf{Effect of Objectness Prior.}
By introducing the objectness prior in the \Obj\ model, pixels spatially and chromatically close to the tree centroid are given higher probability of being that tree. 
Effectively, it expands the learning from only known tree pixels to also include possible pixels. 
However, our task is different from \cite{engineeredObjectness} because many objects are unlabeled, which makes the boundaries generated from instance areas less representative of the background class. 
Our results show that objectness helps when training in a weakly supervised setting with imbalanced data, but the incomplete labeling is not accounted for in this case and the models including our masking regime (\MaskObj\ and \MaskObjThresh) outperform the \Obj\ model.

\textbf{Effect of Mask.}
The masking regime is crucial in an imbalanced and sparsely labeled data setting. 
Without masking, the baseline model failed to predict the target class completely. 
As shown in Table~\ref{tab:ablation}, the mask regime introduced more performance improvements than the objectness prior, with an significantly improved IoU score for models with masking. 
Among those models, the \Mask~model achieved the highest IoU and BA on the delineated data and best recall score on the sparse data. 
In summary, the mask regime is comparatively simple to implement, costs less computational resources, and requires less fine tuning than using the objectness prior.

\textbf{Effect of combining Objectness Prior and Mask.}
The mask regime consistently improved the performance of the models learning from the objectness prior, but there was no improvement compared to the model applying masking without the objectness prior.
A possible reason for this might be that in our complicated urban environment setting, the spatial context of trees might vary quite a bit (e.g., small and large trees) and there is a chance that the objectness map would highlight the object around a tree that is actually not a tree (see Figure~\ref{fig:objectness-map}).
Since the extend and cutoff of the probabilities depend on hyperparameters when creating the prior, there could be settings giving a good performance for one but not all the different scenarios.
This brings us to the conclusion that the objectness prior in its current form does not yield any benefits compared to simply applying the mask regime on its own.

\section{Conclusion}
In this paper, we create a new tree segmentation dataset from public data to train a deep learning model for semantic segmentation of urban trees. 
This dataset is challenging because it consists of incomplete and sparse point labels for trees and carefully selected background objects from OSM.
We address this label incompleteness and sparsity by proposing a loss masking regime into our model design including domain knowledge.
Further, we expand on the weakly supervised technique of learning from objectness priors by utilizing the same masking regime.

Our evaluation shows that the best performance was achieved when only using our masking regime, with a test performance on a fully delineated set of \num{0.43} IoU and \num{84}\% balanced accuracy.
This indicates that the chosen objectness prior was not helpful for this task while our mask regime is beneficial when dealing with incomplete and sparsely annotated data. This even holds when combining it with other approaches, such as the \Obj\ model.
Besides, our mask regime is simple to implement, lower in computational resource requirements, and requires less fine-tuning of hyperparameters. While including \Obj\ or its variants requires to identify and compute an appropriate objectness prior $o$ for each new task and training sample. This overhead is an inhereent aspect of the objectness methods.
Our results on both point labels and a manually delineated evaluation set demonstrates the hidden potential of public datasets for mapping urban trees.

In the future, we will investigate the usefulness of self-supervised networks pretrained on large datasets in conjunction with weak labels to further improve the mapping performance.
Evaluation in other urban areas would also be of interest to validate the generalization performance.
Since our current masking prior is calibrated to our Hamburg dataset, it may not transfer well to new areas. 
Therefore we would like to investigate if the masking prior could be learned or adjusted from unsupervised or semi-supervised networks.

\begin{acks}
This work was supported by the research grant DeReEco from VILLUM FONDEN  (grant number 34306), the PerformLCA project (UCPH Strategic plan 2023 Data+ Pool), and 
the grant ``Risk-assessment of Vector-borne Diseases Based on Deep Learning and Remote Sensing'' (grant number NNF21OC0069116) by the Novo Nordisk Foundation.
\end{acks}

\bibliographystyle{ACM-Reference-Format}
\bibliography{reference}

\end{document}